\documentclass{article}

\usepackage{geometry}

\usepackage[utf8]{inputenc} 
\usepackage[T1]{fontenc}    
\usepackage{hyperref}       
\usepackage{url}            
\usepackage{booktabs}       
\usepackage{amsfonts}       
\usepackage{nicefrac}       
\usepackage{microtype}      
\usepackage{graphicx}
\usepackage{xcolor}
\usepackage{amsmath,amssymb}
\usepackage[ruled,vlined,linesnumbered]{algorithm2e}
\usepackage{footmisc}
\usepackage{mathtools}
\usepackage{bm}
\usepackage{enumitem}
\usepackage{multicol,multirow}
\usepackage{pifont}
\usepackage{subcaption}
\newtheorem{proposition}{Proposition}
\usepackage{mathtools}

\DeclarePairedDelimiterX{\infdivx}[2]{\big(}{\big)}{%
  #1\;\delimsize\|\;#2%
}
\newcommand{\infdiv}{\mathsf{KL}\infdivx}

\newcommand{\Yes}{\textcolor{green}{\ding{51}}}%
\newcommand{\No}{\textcolor{red}{\ding{55}}}%
\DeclareMathOperator{\sign}{sign}
\DeclareMathOperator{\proj}{Proj}

\SetCommentSty{mycommfont}

\title{Improving the Speed and Quality of GAN by Adversarial Training}

\author{%
  Jiachen Zhong \qquad  Xuanqing Liu \qquad Cho-Jui Hsieh \\
  Department of Computer Science \\
  University of California, Los Angeles\\
  Los Angeles CA 90095 \\
  \texttt{\{julightzhong10,xqliu,chohsieh\}@cs.ucla.edu} \\
}

\begin{document}

\maketitle

\begin{abstract}
Generative adversarial networks (GAN) have shown remarkable results in image generation tasks. High fidelity class-conditional GAN methods often rely on stabilization techniques by constraining the global Lipschitz continuity. Such regularization leads to less expressive models and slower convergence speed; other techniques, such as the large batch training, require unconventional computing power and are not widely accessible. 
In this paper, we develop an efficient algorithm, namely FastGAN (Free AdverSarial Training), to improve the speed and quality of GAN training based on the adversarial training technique. We benchmark our method on CIFAR10, a subset of ImageNet, and the full ImageNet datasets. We choose strong baselines such as SNGAN and SAGAN; the results demonstrate that our training algorithm can achieve better generation quality (in terms of the Inception score and Frechet Inception distance) with less overall training time. Most notably, our training algorithm brings ImageNet training to the broader public by requiring 2-4 GPUs.
\end{abstract}

\section{Introduction}
Generating high-quality samples from complex distribution is one of the fundamental problems in machine learning. One of the approaches is by generative adversarial networks (GAN)~\cite{goodfellow2014generative}. Inside the GAN, there is a pair of generator and discriminator trained in some adversarial manner. GAN is recognized as an efficient method to sample from high-dimensional, complex distribution given limited examples, though it is not easy to converge stably. Major failure modes include gradient vanishing and mode collapsing.A long line of research is conducted to combat instability during training, e.g.~\cite{mao2017least,arjovsky2017wasserstein,gulrajani2017improved,miyato2018spectral,zhang2018self,brock2018large}. Notably, WGAN~\cite{arjovsky2017wasserstein} and WGAN-GP~\cite{gulrajani2017improved} propose to replace the KL-divergence metric with Wasserstein distance, which relieve the gradient vanishing problem; SNGAN~\cite{miyato2018spectral} takes a step further by replacing the gradient penalty term with spectral normalization layers, improving both the training stability and sample quality over the prior art. Since then, many successful architectures, including SAGAN~\cite{zhang2018self} and BigGAN~\cite{brock2018large}, utilize the spectral normalization layer to stabilize training. The core idea behind the WGAN or SNGAN is to constrain the smoothness of discriminator. Examples include weight clipping, gradient norm regularization, or normalizing the weight matrix by the largest singular value. However, all of them incur non-negligible computational overhead (higher-order gradient or power iteration).
\par
Apart from spectral normalization, a bundle of other techniques are proposed to aid the training further; among them, essential techniques are two time-scale update rule (TTUR)~\cite{heusel2017gans}, self attention~\cite{zhang2018self} and large batch training~\cite{brock2018large}. However, both of them improve the Inception score(IS)~\cite{salimans2016improved} and Frechet Inception Distance(FID)~\cite{heusel2017gans} at the cost of slowing down the per iteration time or requiring careful tuning of two learning rates. Therefore, it would be useful to seek some alternative ways to reach the same goal but with less overhead. In particular, we answer the following question: 
\begin{quote}\textit{Can we train a plain ResNet generator and discriminator without architectural innovations, spectral normalization, TTUR or big batch size?}
\end{quote}
\par
In this paper, we give an affirmative answer to the question above. We will show that all we need is adding another adversarial training loop to the discriminator. Our contributions can be summarized as follows
\vspace{-5pt}
\begin{itemize}[leftmargin=*,noitemsep]
    \item We show a simple training algorithm that accelerates the training speed and improves the image quality all at the same time.
    \item We show that with the training algorithm, we could match or beat the strong baselines (SNGAN and SAGAN) with plain neural networks (the raw ResNet without spectral normalization or self-attention).
    \item Our algorithm is widely applicable: experimental results show that the algorithm works on from tiny data (CIFAR10) to large data (1000-class ImageNet) using $4$ or less GPUs.
\end{itemize}
\paragraph{Notations} Throughout this paper, we use $G$ to denote the generator network and $D$ to denote the discriminator network (and the parameters therein). $x\in\mathbb{R}^d$ is the data in $d$ dimensional image space; $z$ is Gaussian random variable; $y$ is categorical variable indicating the class id. Subscripts $r$ and $f$ indicates real and fake data. For brevity, we denote the  Euclidean 2-norm as $\|\cdot\|$. $\mathcal{L}$ is the loss function of generator or discriminator depending on the subscripts.
\vspace{-5pt}
\section{Related Work}
\subsection{Efforts to Stabilize GAN Training}\label{GAN_background}
Generative Adversarial Network (GAN) can be seen as a min-max game between two players: the generator ($G$) and discriminator ($D$). At training time, the generator transforms Gaussian noise to synthetic images to fool the discriminator, while the discriminator learns to distinguish the fake data from the real. The loss function is formulated as 
\begin{equation} \label{eq:orig_GAN}
\min_G \max_D \mathop{\mathbb{E}}_{x\sim \mathbb{P}_{\text{real}}}\big[\log D(x)\big]+\mathop{\mathbb{E}}_{x \sim G(z)}\Big[\log\big(1-D(x)\big)\Big],
\end{equation}
where $\mathbb{P}_{\text{real}}$ is the distribution of real data, and $z$ is sampled from the standard normal distribution. Our desired solution of \eqref{eq:orig_GAN} is $G(z)=\mathbb{P}_{\text{real}}$ and $D(x)=0.5$; however, in practice we can rarely get this solution. Early findings show that training $G$ and $D$ with multi-layer CNNs on GAN loss~\eqref{eq:orig_GAN} does not always generate sensible images even on simple CIFAR10 data. Researchers tackle this problem by having better theoretical understandings and intensive empirical experiments. Theoretical works mainly focus on the local convergence to the Nash equilibrium~\cite{mescheder2017numerics,mescheder2018training,wang2019solving} by analyzing the eigenvalues of the Jacobian matrix derived from gradient descent ascent (GDA). Intuitively, if the eigenvalues $\lambda_i$ (maybe complex) lie inside the unit sphere , i.e. $\|\lambda_i\| \le 1$, and the ratio $\frac{Im(\lambda_i)}{Re(\lambda_i)}$ is small, then the convergence to local equilibrium will be fast.
\begin{table}[htb]
    \centering
    \caption{\label{tab:compare-GANs}Comparing the key features and training tricks in GAN models. Our model is a simple combination of ResNet backbone and adversarial training. The model size is calculated for the ImageNet task.}
    \begin{tabular}{ccccc}
    \toprule
      & SNGAN & SAGAN & BigGAN & \textbf{FastGAN (ours)} \\
    \midrule
    ResNet $G/D$ & \Yes & \Yes & \Yes & \Yes \\
    Wider $D$ & \No & \Yes & \Yes & \No \\
    Shared embedding \& skip-z & \No & \No & \Yes & \No \\
    Self attention & \No & \Yes & \Yes & \No\\
    TTUR & \No & \Yes & \Yes & \No\\
    Spectral normalization & \Yes & \Yes & \Yes & \No\\
    Orthogonal regularization & \No & \No & \Yes & \No \\
    Adversarial training & \No & \No & \No & \Yes \\
    Model size & 72.0M & 81.5M & 158.3M & 72.0M \\
    \bottomrule
    \end{tabular}
\end{table}
Empirical findings mainly focus on regularization techniques, evolving from weight clipping~\cite{arjovsky2017wasserstein}, gradient norm regularization~\cite{gulrajani2017improved}, spectral normalization~\cite{miyato2018spectral}, regularized optimal transport~\cite{sanjabi2018convergence} and Lipschitz regularization~\cite{zhou2019lipschitz}, among others. Our method could also be categorized as a regularizer~\cite{DBLP:journals/corr/abs-1906-01527}, in which we train the discriminator to be robust to adversarial perturbation on either real or fake images.

\subsection{Follow the Ridge (FR) algorithm}
A recent correction to the traditional gradient descent ascent (GDA) solver for \eqref{eq:orig_GAN} is proposed~\cite{wang2019solving}. Inside FR algorithm, the authors add a second-order term that drags the iterates back to the ``ridge'' of the loss surface $\mathcal{L}(G, D)$, specifically,
\begin{equation}
    \label{eq:FR-iter}
    \begin{aligned}
    w_G &= w_G-\eta_G\nabla_{w_G}\mathcal{L},\\
    w_D &= w_D+\eta_D\nabla_{w_D}\mathcal{L}+\eta_GH^{-1}_{w_Dw_D}H_{w_Dw_G}\nabla_{w_G}\mathcal{L},
    \end{aligned}
\end{equation}
where we denote $\mathcal{L}$ as the GAN loss in \eqref{eq:orig_GAN}, $H_{w_Dw_D}=\frac{\partial^2\mathcal{L}}{\partial w_D^2}$ and $H_{w_Dw_G}=\frac{\partial^2\mathcal{L}}{\partial w_D\partial w_G}$. With the new update rule, the authors show less rotational behavior in training and an improved convergence to the local minimax. The update rule~\eqref{eq:FR-iter} involves Hessian vector products and Hessian inverse. Comparing with the simple GDA updates, this is computationally expensive even by solving a linear system. Interestingly, under certain assumptions, we can see our adversarial training scheme could as an approximation of \eqref{eq:FR-iter} with no overhead. As a result, our FastGAN converges to the local minimax with fewer iterations (and wall-clock time), even though we dropped the bag of tricks (Table~\ref{tab:compare-GANs}) commonly used in GAN training.
\vspace{-5pt}
\subsection{\label{adv_training}Adversarial training}
Adversarial training~\cite{goodfellow2014explaining} is initially proposed to enhance the robustness of the neural network. The central idea is to find the adversarial examples that maximize the loss and feed them as synthetic training data. Instances of which include the FGSM~\cite{goodfellow2014explaining}, PGD~\cite{madry2017towards}, and free adversarial training~\cite{shafahi2019adversarial}. All of them are approximately solving the following problem
\begin{equation} \label{eq:adv_training}
\begin{split}
\min_{w} \frac{1}{N}\sum_{i=1}^N \Big[\max_{\|\epsilon\|\leq c_{\max}}\mathcal{L}(f(x_i+\epsilon;w),y_i)\Big],
\end{split}
\end{equation}
where $\{(x_i,y_i)\}_{i\in[N]}$ are the training data; $f$ is the model parameterized by $w$; $\mathcal{L}$ is the loss function; $\epsilon$ is the perturbation with norm constraint $c_{\max}$. It is nontrivial to solve \eqref{eq:adv_training} efficiently, previous methods solve it with alternative gradient descent-ascent: on each training batch $(x_i, y_i)$, they launch projected gradient ascent for a few steps and then run one step gradient descent~\cite{madry2017towards}. In this paper, we study how adversarial training helps GAN convergence. To the best of our knowledge, only \cite{zhou2018don} and \cite{liu2019rob} correlate to our idea. However, \cite{zhou2018don} is about unconditional GAN, and none of them can scale to ImageNet. In our experiments, we will include \cite{liu2019rob} as a strong baseline in smaller datasets such as CIFAR10 and subset of ImageNet.
\vspace{-7pt}
\section{FastGAN -- Free AdverSarial Training for GAN}
Our new GAN objective is a three-level min-max-min problem: 
\begin{equation} \label{eq:faster-rob-GAN}
    \min_G \max_D \mathop{\mathbb{E}}_{x\sim \mathbb{P}_{\text{real}}}\big[\min_{\|\epsilon\|\le c_{\max}}\log D(x+\epsilon)\big]+\mathop{\mathbb{E}}_{x \sim G(z)}\Big[\min_{\|\epsilon\|\le c_{\max}}\log\big(1-D(x+\epsilon)\big)\Big].
\end{equation}
Note that this objective function is similar to the one proposed in RobGAN~\cite{liu2019rob}. However, RobGAN cannot scale to large datasets due to the bottleneck introduced by adversarial training (see our experimental results), and no satisfactory explanations are provided to justify why adversarial training improves GAN. 
We made two improvements: First of all, we employ the recent free adversarial training technique~\cite{shafahi2019adversarial}, which simultaneously updates the adversaries and the model weights. Second, in the label-conditioned GAN setting, we improved the loss function used in RobGAN (both are based on cross-entropy rather than hinge loss). 
In the following parts, we first show the connection to follow-the-ridge update rules~\eqref{eq:FR-iter} and then elaborate more on the algorithmic details as well as a better loss function.

\subsection{From \textit{Follow-the-Ridge} (FR) to FastGAN}
Recently \cite{wang2019solving} showed that the Follow-the-Ridge (FR) optimizer improves GAN training, but FR relies on Hessian-vector products and cannot scale to large datasets.
We show that solving \eqref{eq:faster-rob-GAN} by one-step adversarial training can be regarded as an efficient simplification of FR, which partially explains why the proposed algorithm can stabilize and speed up GAN training.
To show this, we first simplify the GAN training problem as follows:
\begin{itemize}[leftmargin=*,noitemsep]
\item Inside the minimax game, we replace the generator network by its output $x_f$. This dramatically simplifies the notations as we no longer need to calculate the gradient through G.
\item The standard GAN loss on fake images~\cite{goodfellow2014generative} is written as $\log\big(1-D(G(z)\big)$. However, we often modify it to $-\log D(G(z))$ to mitigate the gradient vanishing problem.
\end{itemize}
We first consider the original GAN loss:
\begin{equation}
    \label{eq:minmax-problem}
    \min_{G}\max_{D}\mathcal{L}(G, D)
\triangleq\mathop{\mathbb{E}}_{x_r\sim \mathbb{P}_{\text{real}}}\big[\log D(x_r)\big]-\mathop{\mathbb{E}}_{x_f\sim G}\big[\log D(x_f)\big].
\end{equation}
With FR algorithm, the update rule for $D$ (parameterized by $w$) can be written as\footnote{In this equation, we simplify the FR update ($H_{yy}^{-1}$ is dropped). The details are shown in the appendix.}
\begin{equation}
    \label{eq:FR-update}
   w^+ \xleftarrow{\text{FR}} w+\eta_D\Big(\underbrace{\frac{\nabla_wD(x_r)}{D(x_r)}-\frac{\textcolor{blue}{\nabla_w D(x_f)}}{D(x_f)}}_{\text{Gradient ascent}}\Big)-\eta_G\underbrace{H_{wx}\nabla_{x_f}\mathcal{L}(w, x_f)}_{\text{off-the-ridge correction}},
\end{equation}
where $H_{wx}=\frac{\partial^2 \mathcal{L}}{\partial w\partial x_f}$. The last term is regarded as a correction for ``off-the-ridge move'' by $G$-update. If we decompose it further, we will get
\begin{equation}
    \label{eq:FR-correction-term}
    H_{wx}\nabla_{x_f}\mathcal{L}=\frac{\nabla^2_{w, x_f} D(x_f)\nabla_{x_f}D(x_f)}{D(x_f)^2}-\frac{\|\nabla_{x_f} D(x_f)\|^2\textcolor{blue}{\nabla_w D(x_f)}}{D(x_f)^3}.
\end{equation}
Since the first-order term in \eqref{eq:FR-correction-term} is already accounted for in \eqref{eq:FR-update} (both are highlighted in blue), the second-order term in \eqref{eq:FR-correction-term} plays the key role for fast convergence in FR algorithm. However, the algorithm involves Hessian-vector products in each iteration and not very efficient for large data such as ImageNet. Next, we show that our adversarial training could be regarded as a Hessian-free way to perform almost the same functionality. Recall in the adversarial training of GAN, the loss (Eq.~\eqref{eq:faster-rob-GAN}) becomes (fixing $c_{\max}=1$ for brevity)
\begin{equation}
    \label{eq:RobGAN}
    \begin{aligned}
\mathcal{L}^{\text{adv}}(G, D)
&\triangleq\mathop{\mathbb{E}}_{x_r\sim \mathbb{P}_{\text{real}}}\big[\min_{\|\epsilon\|_2\le 1}\log D(x_r+\epsilon)\big]-\mathop{\mathbb{E}}_{x_f\sim G}\max_{\|\epsilon\|_2\le 1}\log D(x_f+\epsilon)\\
&\approx \mathop{\mathbb{E}}_{x_r\sim\mathbb{P}_{\text{real}}}\Big[\log D(x_r)-\frac{\|\nabla_{x_r}D(x_r)\|}{D(x_r)}\Big]-\mathop{\mathbb{E}}_{x_f\sim G}\Big[\log D(x_f)+\frac{\|\nabla_{x_f}D(x_f)\|}{D(x_f)}\Big],
    \end{aligned}
\end{equation}
here we assume the inner minimizer only conducts one gradient descent update (similar to the algorithm proposed in the next section), and use first-order Taylor expansion to approximate the inner minimization problem.
So the gradient descent/ascent updates will be
\begin{equation}
\resizebox{.5\linewidth}{!}{$
    w^+ \xleftarrow{\text{GDA}} w+\eta_D\Big(\underbrace{\frac{\nabla_wD(x_r)}{D(x_r)}-\frac{\nabla_w D(x_f)}{D(x_f)}}_{\text{Gradient ascent}}\Big)-\overrightarrow{\bm{\bigtriangledown}}$.
}
\end{equation}
Here we define $\overrightarrow{\bm{\bigtriangledown}}$ as the gradient correction term introduced by adversarial training, i.e.
\begin{equation}
    \label{eq:gradient-correction-from-ADV}
    \resizebox{.93\linewidth}{!}{$
    \overrightarrow{\bm{\bigtriangledown}}=\sum\limits_{x\in\{x_r, x_f\}}\nabla_w\Big(\frac{\|\nabla_{x}D(x)\|}{D(x)}\Big)=\sum\limits_{x\in\{x_r, x_f\}}\frac{\nabla^2_{w,x}D(x)\nabla_{x}D(x)}{\|\nabla_{x}D(x)\|D(x)}-\frac{\|\nabla_{x}D(x)\|^2\nabla_wD(x)}{D(x)^2}.$}
\end{equation}
Comparing~\eqref{eq:gradient-correction-from-ADV} with~\eqref{eq:FR-correction-term}, we can see both of them are a linear combination of second-order term and first-order term, except that the FR is calculated on fake images $x_f$ while our adversarial training can be done on $x_r$ or $x_f$. The two becomes the same up to a scalar $D(x)$ if 1-Lipschitz constraint is enforced on the discriminator $D$, in which case $\|\nabla_x D(x)\|=1$. The constraint is commonly seen in previous works, including WGAN, WGAN-GP, SNGAN, SAGAN, BigGAN, etc. However, we do not put this constraint explicitly in our FastGAN.

\subsection{Training algorithm}
Our algorithm is described in Algorithm~\ref{alg:fastrobgan}, the main difference from classic GAN training is the extra for-loop inside in updating the discriminator. In each iteration, we do $\texttt{MAX\_ADV\_STEP}$ steps of adversarial training to the discriminator network. Inside the adversarial training loop, the gradients over input images (denoted as $g_x$) and over the discriminator weights (denoted as $g_{w_D}$) can be obtained with just one backpropagation. We train the discriminator with fake data $(x_f, y_f)$ immediately after one adversarial training step. A handy trick is applied by reusing the same fake images (generated at the beginning of each $\texttt{D\_step}$) multiple times -- we found no performance degradation and faster wall-clock time by avoiding some unnecessary propagations through $G$.
\begin{algorithm}[htb]
\DontPrintSemicolon
\SetAlgoLined
\caption{\label{alg:fastrobgan}FastGAN training algorithm}
\KwData{Training set $\{(x_i,y_i)\}$, max perturbation $c_{\max}$, learning rate $\eta$.}
Initialize: generator $G$, discriminator $D$, and perturbation $\epsilon$.\\
\While{not converged}{
  \tcc{Train discriminator for MAX\_D\_STEP steps}
\For{\texttt{D\_step=1..MAX\_D\_STEP}}{
\tcc{Prepare real and fake data, same data will be used multiple times in the nested for-loop.}
$z\gets \mathcal{N}(0, 1)$; $y_f\gets \texttt{Categorical}(1..C)$\\
$x_f \gets G(z, y_f)$ \\
$x_r, y_r \gets \texttt{sample}(\{x_i,y_i\}, i\in[N])$ \\

\For{\texttt{adv\_step=1..MAX\_ADV\_STEP}}{
\tcc{Conduct free adversarial training on real data}
$g_{x},\ g_{w_D} \gets \nabla_{x_{r}} \mathcal{L}\big(D(x_r+\epsilon),y_r\big)$, $\nabla_{w_D} \mathcal{L}\big(D(x_r+\epsilon),y_r\big)$\\
$w_D \gets w_D + \eta \cdot g_{w_D}$\\
$\epsilon \gets \proj_{\|\cdot\|\le c_{\max}}\big(\epsilon - c_{\max} \cdot \sign(g_{x})\big)$\\
\tcc{Reuse the fake data during adversarial training for best speed}
$g_{w_D}\gets \nabla_{w_D}\mathcal{L}\big(D(x_f), y_f)\big)$\\
$w_D\gets w_D+\eta\cdot g_{w_D}$\\
}
}
\tcc{Train generator for one step}
$z\gets \mathcal{N}(0, 1)$; $y_f\gets \texttt{Categorical}(1..C)$\\
$x_{f} \gets G(z, y_f)$ \\
$g_{w_G} \gets \nabla_{w_G}\mathcal{L}\big(D(x_f),y_f\big)$ \\
$w_G \gets w_G - \eta \cdot g_{w_G}$
}
\end{algorithm}

Contrary to the common beliefs~\cite{heusel2017gans,zhang2018self} that it would be beneficial (for stability, performance and convergence rate, etc.) to have different learning rates for generator and discriminator, our empirical results show that once the discriminator undergoes robust training, it is no longer necessary to tune learning rates for two networks. 
\subsection{\label{sec:improved_objective_loss}Improved loss function}
Projection-based loss function~\cite{miyato2018cgans} is dominating current state-of-the-art GANs (e.g., \cite{wu2019logan,brock2018large}) considering the stability in training; nevertheless, it does not imply that traditional cross-entropy loss is an inferior choice. In parallel to other works~\cite{gong2019twin}, we believe ACGAN loss can be as good as projection loss after slight modifications. First of all, consider the objective of discriminator $\max_D \mathcal{L}_D\coloneqq\mathcal{L}_D^r+\mathcal{L}_D^f$ where $\mathcal{L}_D^r$ and $\mathcal{L}_D^f$ are the likelihoods on real and fake minibatch, respectively. For instance, in ACGAN we have
\begin{equation}
    \label{eq:AC-GAN-loss}
        \mathcal{L}_D^r=\mathbb{E}_{(x_r, y_r)\sim \mathbb{P}_{\text{real}}}\big[\log \Pr\big(\text{real}\wedge y_r|D(x_r)\big)\big],   \ \ \ 
        \mathcal{L}_D^f=\mathbb{E}_{(x_f, y_f)\sim G}\big[\log \Pr\big(\text{fake}\wedge y_f|D(x_f)\big)\big].
\end{equation}
We remark that in the class-conditioning case, the discriminator contains two output branches: one is for binary classification of real or fake, the other is for multi-class classification of different class labels. The log-likelihood should be interpreted under the joint distribution of the two. However, as pointed out in~\cite{gong2019twin,liu2019rob}, the loss~\eqref{eq:AC-GAN-loss} encourages a degenerate solution featuring a mode collapse behavior. The solution of TwinGAN~\cite{gong2019twin} is to incorporate another classifier (namely ``twin-classifier'') to help generator $G$ promoting its diversity; while RobGAN~\cite{liu2019rob} removes the classification branch on fake data:
\begin{equation}
    \label{eq:RobGAN-loss}
    \mathcal{L}_D^f=\mathbb{E}_{(x_f,y_f)\sim G}\big[\log \Pr\big(\text{fake}|D(x_f)\big)\big].
\end{equation}

Overall, our FastGAN uses a similar loss function as RobGAN~\eqref{eq:RobGAN-loss}, except we changed the adversarial part from probability to hinge loss~\cite{lim2017geometric,miyato2018spectral}, which reduces the gradient vanishing and instability problem. As to the class-conditional branch, FastGAN inherits the auxiliary classifier from ACGAN as it is more suitable for adversarial training. However, as reported in prior works~\cite{odena2017conditional,miyato2018cgans}, training a GAN with auxiliary classification loss 
has no good intra-class diversity. To tackle this problem, we added a KL term to $\mathcal{L}_D^f$. Therefore, $\mathcal{L}_D^r$ and $\mathcal{L}_D^f$ become:
\begin{equation}
    \label{eq:FastGAN-Ld}
    \begin{aligned}
    &\mathcal{L}_D^r=\mathbb{E}_{(x_r,y_r)\sim \mathbb{P}_\text{real}}\big[-\max\big(0, 1-D(x_r)\big)+\Pr\big(y_r|D(x_r)\big)\big],\\
    &\mathcal{L}_D^f=\mathbb{E}_{(x_f,y_f)\sim G}\big[-\max\big(0, 1+D(x_f)\big)\big]-\alpha_c^f\cdot\mathsf{KL}(\Pr(y_f|D(x_f)), \mathcal{U})\big].
    \end{aligned}
\end{equation}
where $\alpha_c^f\in(0, 1)$ is a coefficient, $\mathcal{U}=(1/C, 1/C, \dots, 1/C)^T$ is a uniform categorical distribution among all $C$-classes. Our loss~\eqref{eq:FastGAN-Ld} is in sharp difference to ACGAN loss in~\eqref{eq:AC-GAN-loss}: in ACGAN, the discriminator gets rewarded by assigning high probability to $y_f$; while our FastGAN is encouraged to assign a uniform probability to all labels in order to enhance the intra-class diversity of the generated samples. Additionally, we found that it is worth to add another factor $\alpha_c^g$ to $\mathcal{L}_G$ to balance between image diversity and fidelity, so the generator loss in FastGAN is defined as
\begin{equation}
    \label{eq:FastGAN-Lg}
     \mathcal{L}_G = \mathbb{E}_{(x_f,y_f)\sim G}\big[-\mathop{\mathbb{E}}[ D(x_f)] - \alpha_c^g\cdot\log \Pr(y_f|D(x_f))\big].
\end{equation}
Overall, in the improved objectives of FastGAN, $G$ is trained to minimize  $\mathcal{L}_G$ while $D$ is trained to maximize $\mathcal{L}_D^r+\mathcal{L}_D^f$.
\vspace{-5pt}
\section{\label{experiments}Experiments}
\vspace{-5pt}
In this section, we test the performance of FastGAN on a variety of datasets. For the baselines, we choose SNGAN~\cite{miyato2018spectral} with projection discriminator, RobGAN~\cite{liu2019rob}, and SAGAN~\cite{zhang2018self}. Although better GAN models could be found, such as the BigGAN~\cite{brock2018large} and the LOGAN~\cite{wu2019logan}, we do not include them because they require large batch training (batch size~$>$~1k) on much larger networks (see Table~\ref{tab:compare-GANs} for model sizes). 

\textbf{Datasets.}\quad We test on following datasets: CIFAR10, CIFAR100~\cite{krizhevsky2009learning}, ImageNet-143~\cite{miyato2018spectral,liu2019rob}, and the full ImageNet~\cite{russakovsky2015imagenet}. Notablly the ImageNet-143 dataset is an $143$-class subset of ImageNet~\cite{russakovsky2015imagenet}, first seen in SNGAN~\cite{miyato2018spectral}. We use both 64x64 and 128x128 resolutions in our expriments.

\textbf{Choice of architecture.}\quad As our focus is not on architectural innovations, for a fair comparison, we did the layer-by-layer copy of the ResNet backbone from SNGAN (spectral normalization layers are removed). So our FastGAN, SNGAN, and RobGAN are directly comparable, whereas SAGAN is bigger in model size. For experiments on CIFAR, we follow the architecture in WGAN-GP~\cite{gulrajani2017improved}, which is also used in SNGAN~\cite{miyato2018spectral,miyato2018cgans}.

\textbf{Optimizer.}\quad We use Adam~\cite{kingma2014adam} with learning rate $\eta_0=0.0002$ and momentum $\beta_1=0.0$, $\beta_2=0.9$ (CIFAR/ImageNet-143) or $\beta_2=0.99$ (ImageNet) for both $D$ and $G$. We use exponential decaying learning rate scheduler: $\eta_t = \eta_0 \cdot e^{t/\kappa }$
where $t$ is the iteration number. 

Other hyperparameters are attached in appendix.
\subsection{CIFAR10 and CIFAR100}
\begin{figure}
\begin{center}
\includegraphics[width=1.0\linewidth]{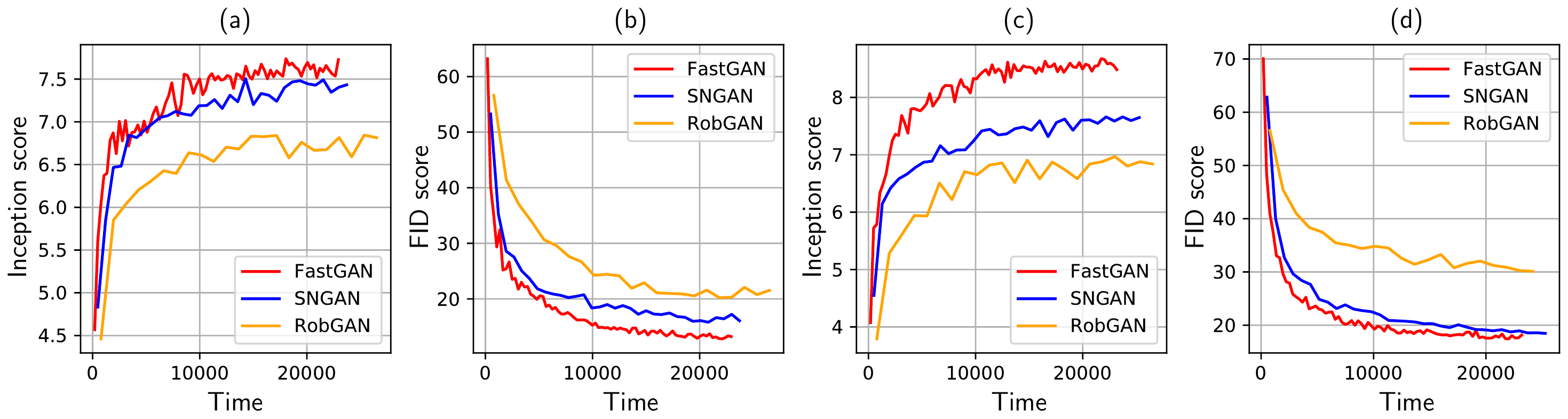}
\end{center}
\vspace{-7pt}
   \caption{Comparing the convergence of GANs on CIFAR10 (a, b) and CIFAR100 (c, d) datasets.}
\label{fig:conv_cifar}
\end{figure}

We report the experimental results in Table~\ref{tab:cifar-exp}. We remind that SAGAN does not contain official results on CIFAR, so we exclude it form this experiment. The metrics are measured after all GANs stop improving, which took $\sim 2.5\times 10^4$ seconds. As we can see, our FastGAN is better than SNGAN and RobGAN at CIFAR dataset in terms of both IS score and FID score. Furthermore, to compare the convergence speed, we exhibit the learning curves of all results in Figure~\ref{fig:conv_cifar}. From this figure, we can observe a consistent acceleration effect from FastGAN.
\begin{table}[htb]
  \caption{\label{tab:cifar-exp}Results on CIFAR10 and CIFAR100. IS indicates the inception score (higher better), FID is the Fréchet Inception Distance (lower better). Time is measured in seconds. Wining results are displayed in \textbf{bold}, collapsed results are displayed in \textcolor{cyan}{blue}.}
  \centering
  \scalebox{0.9}{
  \begin{tabular}{lcccccc}
    \toprule
    & \multicolumn{3}{c}{CIFAR10} & \multicolumn{3}{c}{CIFAR100}  \\
     \cmidrule(r){2-4} \cmidrule(l){5-7}
         & IS $\uparrow$ & FID $\downarrow$ & Time $\downarrow$  & IS $\uparrow$ & FID $\downarrow$ & Time $\downarrow$ \\
    \midrule
    Real data & $11.24\pm.10$ & $5.30$ & -- & $14.79\pm .15$ & $5.91$ & -- \\
    \midrule
    SNGAN & $7.47\pm.13$ & $14.59$ & $2.45\times 10^4$ & $7.86\pm .21$ & $18.29$  & $2.56\times 10^4$ \\
    RobGAN & $6.95\pm .06$ & $20.75$ & $2.50\times 10^4$ & $7.24\pm.21$  & $28.27$ & $2.6\times 10^4$ \\ 
    FastGAN &  $\bm{7.76\pm .12}$ &  $\bm{12.97}$ &  $\bm{2.28\times 10^4}$ &  $\bm{8.87\pm .06}$ & $\bm{17.27}$  & $\bm{2.31\times 10^4}$\\
    \bottomrule
    + Revert to RobGAN loss & $7.14\pm.18$  &  $20.90$  & $2.28\times 10^4$  & $\textcolor{cyan}{6.26\pm .19}$  & $\textcolor{cyan}{43.69}$   & $\textcolor{cyan}{2.31\times 10^4}$\\
    + Disable adv. training &  $7.79\pm .12$  & $13.93$  &  $2.28\times 10^4$ & $8.38\pm .21$   & $19.43$  & $2.31\times 10^4$\\
    + Constant lr. &   $\bm{8.09\pm.11}$ &  $14.67$ &  $2.28\times 10^4$  &  $\bm{8.95\pm.31}$ & $19.53$ & $2.31\times 10^4$\\
    + Disable KL-term~\eqref{eq:FastGAN-Ld} &  $7.37\pm .17$ & $16.26$  &  $2.28\times 10^4$ & $\textcolor{cyan}{7.42\pm.18}$  & $\textcolor{cyan}{24.02}$ & $\textcolor{cyan}{2.31\times 10^4}$\\
    \bottomrule
  \end{tabular}}
\end{table}
\par
\begin{figure}[htb]
    \centering
    \includegraphics[width=0.9\linewidth]{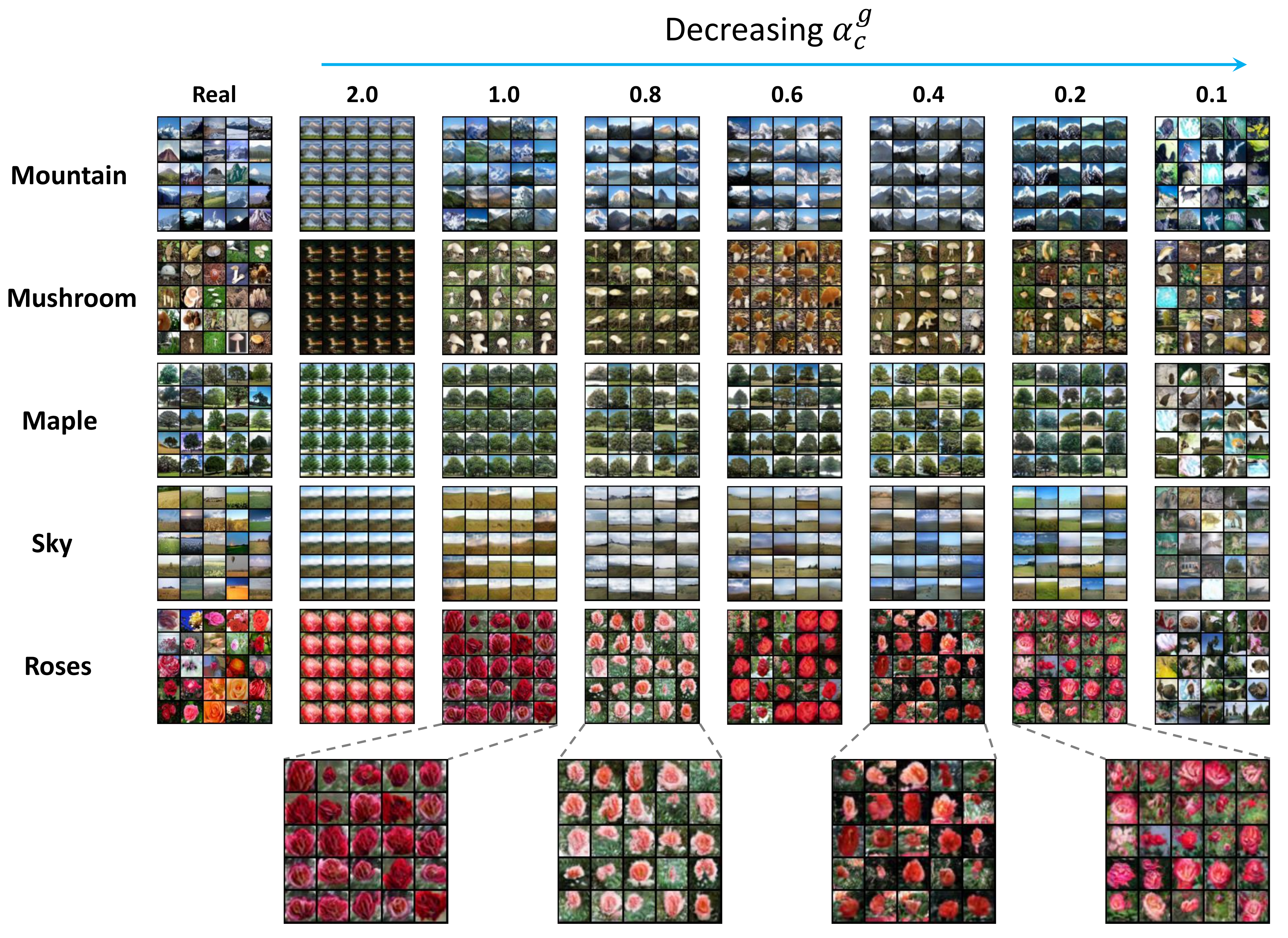}
    \caption{The effect of $\alpha_c^g$ on the diversity - fidelity tradeoff in CIFAR100 experiments. Specifically in the last row, when $\alpha_c^g\in [0.8, 1.0]$ the distributions are unimodal (roses all in the same color: either dark red or peach); when $\alpha_c^g\in [0.2, 0.4]$, distributions are multimodal so we could see roses of more colors. If the coefficient is as small as $0.1$, the distributions become more complex, but we could no longer recognize the images.}
    \label{fig:alpha_cg}
\end{figure}

Next, we study which parts of our FastGAN attribute to performance improvement. To this end, we disable some essential modifications of FastGAN (last four rows in Table~\ref{tab:cifar-exp}). We also try different $\alpha_c^g$ in the experiments of CIFAR100 in Figure~\ref{fig:alpha_cg} and show it effectively controls the tradeoff between diversity and fidelity.
\subsection{ImageNet-143}
\begin{figure}[htb]
\begin{center}
\includegraphics[width=1.0\linewidth]{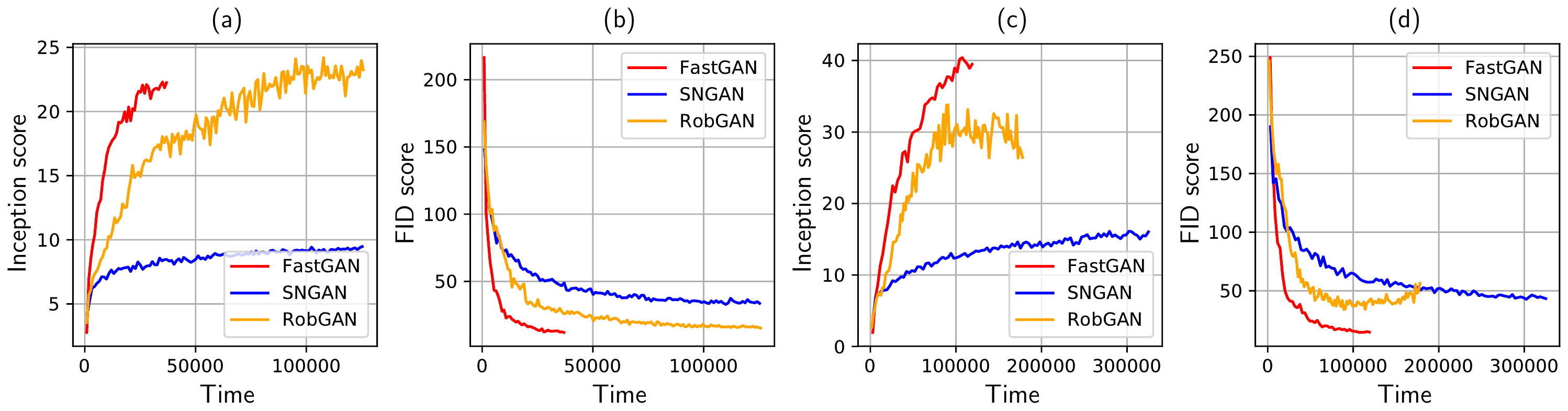}
\end{center}
\vspace{-7pt}
   \caption{Comparing the convergence of GANs on 64x64 resolution (in (a) and (b)) and 128x128 resolution (in (c) and (d)) of the ImageNet-143 dataset.}
\label{fig:conv_catdog}
\end{figure}

ImageNet-143 is a dataset first appeared in SNGAN. As a test suite, the distribution complexity stands between CIFAR and full ImageNet as it has 143 classes and 64x64 or 128x128 pixels. We perform experiments on this dataset in a similar way as in CIFAR. The experimental results are shown in Table~\ref{Tab:imgnet143gq}, in which we can see FastGAN overtakes SNGAN and RobGAN in both IS and FID metrics. As to the convergence rate, as shown in Figure~\ref{fig:conv_catdog}, FastGAN often requires $2{\sim}3$ times less training time compared with SNGAN and RobGAN to achieve the same scores.
 \begin{table}[htb]
  \caption{Results on ImageNet-143.}
  \label{Tab:imgnet143gq}
  \centering
  \resizebox{0.95\textwidth}{!}{
  \begin{tabular}{lcccccc}
    \toprule
    & \multicolumn{3}{c}{64x64 pixels} & \multicolumn{3}{c}{128x128 pixels}  \\
    \cmidrule(r){2-4}\cmidrule(l){5-7}
 & IS $\uparrow$ & FID $\downarrow$ & Time $\downarrow$  & IS $\uparrow$ & FID $\downarrow$ & Time $\downarrow$ \\
    \midrule
Real data & $27.9\pm 0.42$  & $0.48$ & -- & $53.01\pm 0.56$  & $0.42$ & -- \\
\midrule
SNGAN & $10.70\pm 0.17$ &  $29.70$ & $2.2\times 10^5$ & $26.12\pm 0.29$ &  $30.83$ & $5.9\times 10^5$ \\
RobGAN & $24.62\pm0.31$ & $14.64$ & $1.8\times 10^5$ & $33.81\pm 0.47$ & $33.98$ & $1.8\times 10^5$ \\
FastGAN & $22.23\pm 0.35$ & $\bm{12.04}$  &  $\bm{3.6\times 10^4}$ & $40.41\pm 0.49$ &  $\bm{14.48}$ & $\bm{7.1\times 10^4}$ \\
\midrule
+ Revert to RobGAN loss & $25.49\pm0.29$ &  $14.03$ &  $3.6\times 10^4$ & $\bm{45.94\pm0.52}$ &  $25.38$ & $7.1\times 10^4$\\
+ Disable KL-term~\eqref{eq:FastGAN-Ld} & $\bm{25.61\pm0.47}$ &  $13.94$ &  $3.6\times 10^4$ & $42.54\pm 0.40$ &  $17.51$ & $7.1\times 10^4$ \\
    \bottomrule
  \end{tabular}
  }
\end{table}
\subsection{ImageNet}
\begin{table}[h!]
  \caption{Results on full ImageNet.}
  \label{Tab:imgnetgq}
  \centering
  \scalebox{0.83}{
  \begin{tabular}{lccc|lccc}
    \toprule
         & IS $\uparrow$ & FID $\downarrow$ & Time $\downarrow$ &  & IS $\uparrow$ & FID $\downarrow$ & Time $\downarrow$ \\
    \midrule
Real data & $217.43\pm 5.17$  & $1.55$ & -- & & & \\
\midrule
\multicolumn{4}{l}{\textit{Trained with batch size $=64$}} & \multicolumn{4}{l}{\textit{Trained with batch size $=256$}}\\
SNGAN & $36.8$  &  $27.62$ & $2.98\times 10^6$ & SAGAN & $52.52$ &  \bm{$18.65$} & $1.79\times 10^6$ \\
FastGAN & \bm{$51.43\pm 2.51$} & \bm{$22.60$}  &  \bm{$1.34\times 10^6$} &  FastGAN & \bm{$65.57\pm2.65$} &  $19.41$ &  \bm{$1.10\times 10^6$}\\
    \bottomrule
  \end{tabular}}
\end{table}
In this experiment, we set SNGAN and SAGAN as the baselines. RobGAN is not included because no ImageNet experiment is seen in the original paper, nor can we scale RobGAN to ImageNet with the official implementation. The images are scaled and cropped to 128x128 pixels. A notable fact is that SNGAN is trained with batch size $=64$, while SAGAN is trained with batch size $=256$. To make a fair comparison, we train our FastGAN with both batch sizes and compare them with the corresponding official results. From Table~\ref{Tab:imgnetgq}, we can generally find the FastGAN better at both metrics, only the FID score is slightly worse than SAGAN. Considering that our model contains no self-attention block, the FastGAN is $13\%$ smaller and $57\%$ faster than SAGAN in training.
\vspace{-10pt}
\section{Conclusion}
\vspace{-10pt}
In this work, we propose the FastGAN, which incorporates free adversarial training strategies to reduce the overall training time of GAN. Furthermore, we further modify the loss function to improve generation quality. We test FastGAN from small to large scale datasets and compare it with strong prior works. Our FastGAN demonstrates better generation quality with faster convergence speed in most cases.


\appendix
\section{Local stability of the simplified \textit{Follow-the-Ridge} updates}
We justify the stability of our simplified FR update rule in Eq. (6), i.e.
\begin{equation}
    \label{eq:simplified-FR}
    \begin{aligned}
    x_{t+1}&=x_t-\eta_x\nabla_x f(x_t, y_t),\\
    y_{t+1}&=y_t+\eta_y\nabla_y f(x_t, y_t)-\eta_xH_{yx}\nabla_x f(x_t, y_t).
    \end{aligned}
\end{equation}
Compared with the original update rule in \cite{wang2019solving}, our update rule is essentially setting $H_{yy}=-I$. Similar to \cite{wang2019solving}, we analyze the spectral norm of Jacobian update. Before that, we consider the Jacobian at local Nash Equilibrium $(x^*, y^*)$ (note that \cite{wang2019solving} assumes local minimax). From~\cite{jin2019local}, we have following properties under local Nash Equilibrium
\begin{proposition}[First order necessary condition]
Assuming f is differentiable, any local Nash equilibrium
satisfies $\nabla_x f=0$ and $\nabla_y f=0$, where $f$ is the minimax objective function.
\label{prop:first-order-Nash}
\end{proposition}
\begin{proposition}[Second order necessary condition]
\label{prop:second-order-Nash}
Assuming f is twice-differentiable, any local Nash
equilibrium satisfies $\nabla_{xx}^2 f\preceq 0$ and $\nabla_{yy}^2 f \succeq 0$.
\end{proposition}
Based on the properties above, it becomes straightforward to have following deductions
\begin{equation}
    J\triangleq \begin{pmatrix} \frac{\partial x_{t+1}}{\partial x_t} & \frac{\partial x_{t+1}}{\partial y_t} \\ \frac{\partial y_{t+1}}{\partial x_t} & \frac{\partial y_{t+1}}{\partial y_t}\end{pmatrix}= I-\eta_x \begin{pmatrix} I & \\ H_{yx} & I\end{pmatrix} \begin{pmatrix} H_{xx} & H_{xy}\\ -cH_{yx} & -cH_{yy}\end{pmatrix},
\end{equation}
where $c=\eta_y/\eta_x$. As in~\cite{wang2019solving} we have following similar transformation
\begin{equation}
    \begin{pmatrix}I & \\ -H_{yx} & I\end{pmatrix}J\begin{pmatrix} I & \\ H_{yx} & I \end{pmatrix}=I-\eta_x\begin{pmatrix}
    H_{xx}+H_{xy}H_{yx} & H_{xy}\\ & cI
    \end{pmatrix}
\end{equation}
with a small enough $\eta_x$, we can see the Jacobian matrix $J$ always have specral radius $\rho(J)<1$ given the positive definitive $H_{xx}$, $H_{xy}H_{yx}$ and $cI$.
\section{Experimental settings}
\subsection{Training hyperparameters}
Throughout all experiments, we set the steps of updating $D$ per iteration $MAX\_D\_STEP=1$ and the steps of free adversarial training $MAX\_ADV\_STEP=2$ in our Algorithm~1. Other hyperparameters are searched manually, we list them in Table~\ref{Tab:hypparam}.
 \begin{table}[htb]
  \caption{Hyperparameters settings of FastGAN training}
  \label{Tab:hypparam}
  \centering
   \begin{minipage}{\linewidth}
  \footnotesize
  \begin{tabular}{lccccccc}
    \toprule
    & total iter. ($\times10^{3}$) & batch size  & $e$\footnote{$e$ is the rate of exponential decay, i.e. $\eta_t=\eta_0\cdot 0.5^{t/\kappa}$} & $\kappa$ ($\times10^{3}$) & $c_{\max}$  & $\alpha_c^g$ & $\alpha_c^f$ \\
    \midrule
    CIFAR10  &  $240$ & 64 &  0.5 & 80 & $6\times10^{-3}$ &  1.0 &  1.0\\
    CIFAR100 &  $240$ & 64  &  0.5 & 80  & $6\times10^{-3}$ & 0.2 &  1.0\\
    IMAGENET-143 (64px)  &  $120$ & 64   & 0.5 & 30 & $3\times10^{-2}$ & 1.0 &  1.0\\
    IMAGENET-143 (128px)  &  $120$ & 64  & 0.5 & 30  & $3\times10^{-2}$ & 1.0 &  1.0 \\
    IMAGENET  &  $1200$ & 64  & 0.5 & 500  & $1\times10^{-2}$ & 0.2 &  0.05 \\
    IMAGENET  &  $650$  & 256  & 0.5& 250  & $1\times10^{-2}$ & 0.2 &  0.05 \\
    \bottomrule
  \end{tabular}
  \end{minipage}
\end{table}

\subsection{Datasets and pre-processing steps}
We tried 4 datasets in our experiments: CIFAR10, CIFAR100, ImageNet-143, and the full ImageNet. \textbf{CIFAR10} and \textbf{CIFAR100} both contain 50,000 32$\times$32 resolution RGB images in 10 and 100 classes respectively. \textbf{ImageNet-143} is a subset of ImageNet which contains 180,373 RGB images in 143 classes. The experiments were conducted on 64$\times$64 and 128$\times$128 resolutions. The full \textbf{ImageNet} contains 1,281,167 RGB images in 1000 classes. Each class contains approximately 1300 samples. Our experiments ran on 128$\times$128 resolution.
\par
Throughout all experiments, we scale all the values of images to $[-1, 1]$. In addition, during the training, we perform random horizontal flip and random cropping and resizing, scaling from 0.8 to 1.0, on the training images.

\subsection{Evaluation metrics}
We choose Inception Score (IS)~\cite{salimans2016improved} and Frechet Inception distance (FID)~\cite{heusel2017gans} to assess the quality of fake images. 

IS utilizes the pre-trained Inception-v3 network to measure the KL divergence between the conditional class distribution and marginal class distribution:
\begin{equation}\label{eq:IS}
\begin{aligned}
    \text{IS} & = \exp\Big(\frac{1}{N}\sum^N_{n=1}\big[\infdiv{p(y|x_n)}{p(y)} \big]\Big)
\end{aligned}
\end{equation}
where $x_n\sim G$ are samples for testing, the conditional class distribution $p(y|x)$ is given by the pre-trained Inception-v3 networks, and the marginal class distribution $p(y)$ is calculated by using $\frac{1}{N}\sum^N_{n=1}p(y|x_n)$. In this work, we use 50k samples to measure the IS for all experiments including the IS score of real training data.

FID measures the 2-Wasserstein distance between two groups of samples drawn from distribution $p_1$ and $p_2$ in the feature space of Inception-v3 network:
\begin{equation}\label{eq:FID}
\begin{aligned}
    \text{FID} & = \|\mu_{p_1}-\mu_{p_2}\|^2_2 + \mathsf{Tr}(\Sigma_{p_1}+\Sigma_{p_2} - 2(\Sigma_{p_1}\Sigma_{p_2})^{\frac{1}{2}})
\end{aligned}
\end{equation}
where $(\mu_{p_1}, \Sigma_{p_1})$, $(\mu_{p_2}, \Sigma_{p_2})$ are the means and covariance matrices in activations. Here $p_1$ is the real distribution and $p_2$ is the generated distribution. We use the whole dataset to calculate $(\mu_{p_1}, \Sigma_{p_1})$. For the generated samples, we randomly sample 5k data in CIFAR experiments and 50k in other experiments.

\subsection{Training speed comparison}
We ran all experiments with no more than four Nvidia 1080 Ti GPUs, depending on the datasets. We also repeat measurements multiple times but it does not show much variations. We report the details in Table~\ref{Tab:speed}.
 \begin{table}
  \caption{Speed of different experiments}
  \label{Tab:speed}
  \centering
  \resizebox{0.9\textwidth}{!}{
  \begin{tabular}{l|cccc}
    \toprule
 & Method  &  Speed  & Total iter. & Total time \\
 &    & (Sec./$10^3$ iter.)  & ($\times 10^3$) & (seconds) \\
    \midrule
CIFAR10 & FastGAN  &  95.5$\pm$1.6 & 240  & $2.29\times 10^4$  \\
(1 GPU) & SNGAN  &  245.1$\pm$2.4 & 100  & $2.45\times 10^4$  \\
& RobGAN  & 385.1$\pm$3.1 & 65  & $2.50\times 10^4$  \\
    \midrule
CIFAR100 & FastGAN  &  96.4$\pm$0.9 & 240  & $2.31\times 10^4$  \\
(1 GPU) & SNGAN  &  255.7$\pm$1.2 & 100  & $2.56\times 10^4$  \\
& RobGAN  &  397.1$\pm$1.1 & 65  & $2.60\times 10^4$ \\
\midrule
ImageNet-143 64px & FastGAN  &  301.3$\pm$1.4 & 120  & $3.61\times 10^4$  \\
(1 GPU) & SNGAN  &  741.2$\pm$2.1 & 300  & $2.22\times 10^5$   \\
& RobGAN  &  1496.7$\pm$4.1 & 120 &  $1.90\times 10^5$  \\
\midrule
ImageNet-143 128px  & FastGAN  &  595.3$\pm$2.4 & 120  & $7.14\times 10^4$  \\
(2 GPUs)  & SNGAN  &  1310.2$\pm$5.9 & 450 & $5.89\times 10^5$  \\
& RobGAN  &  2993.2$\pm$5.5 &  60 & $1.80\times 10^5$ \\
\midrule
ImageNet BS=64 & FastGAN  &  1113.2$\pm$5.1 & 1200  & $1.36\times 10^6$  \\
(2 GPUs)  & SNGAN  &  3506.2$\pm$4.9 & 850  & $2.98\times 10^6$  \\
\midrule
ImageNet BS=256 & FastGAN  &  1699.6$\pm$4.5 & 650  & $1.10\times 10^6$  \\
(4 GPUs) & SAGAN  &  1793.3$\pm$6.5 & 1000  & $1.79\times 10^6$ \\
\midrule
    \bottomrule
  \end{tabular}
  }
\end{table}
For RobGAN on CIFAR, we set the number of $D$ updates per iteration to 5, the adversarial training steps are 3, the $L_{\infty}$ PGD bound is 0.006, and PGD step size is 0.002.

\section{More ablation studies}
\subsection{Sensitivity on $c_{\max}$}
\begin{figure}[!htbp]
\begin{subfigure}{.5\textwidth}
  \centering
  \includegraphics[width=.9\linewidth]{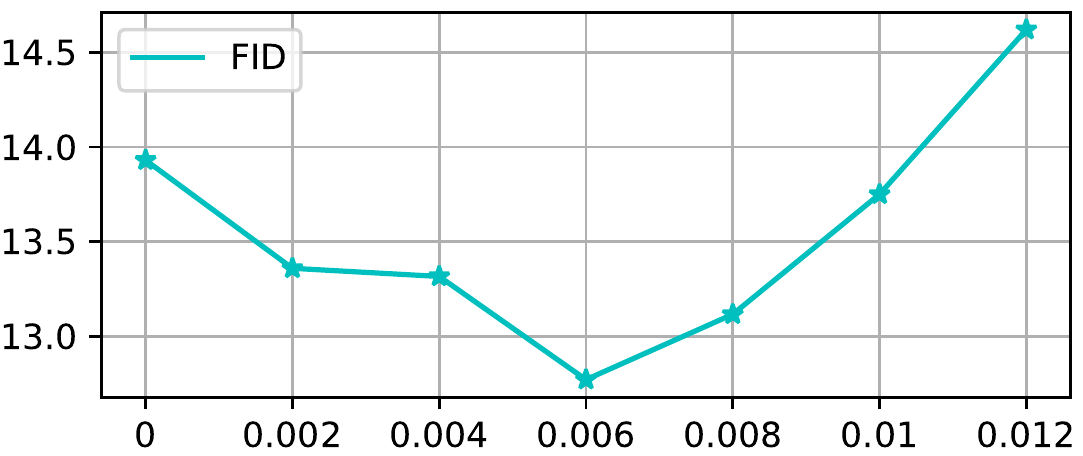}
  \caption{FID on CIFAR10}
  \label{fig:cmax_cifar10_FID}
\end{subfigure}%
\begin{subfigure}{.5\textwidth}
  \centering
  \includegraphics[width=.9\linewidth]{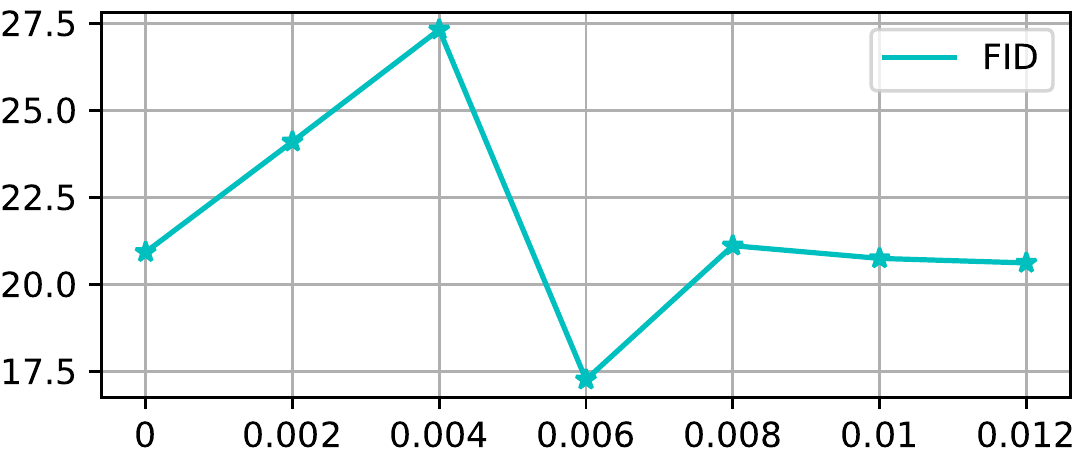}
  \caption{FID on CIFAR100}
  \label{fig:cmax_cifar100_FID}
\end{subfigure}
\begin{subfigure}{.5\textwidth}
  \centering
  \includegraphics[width=.9\linewidth]{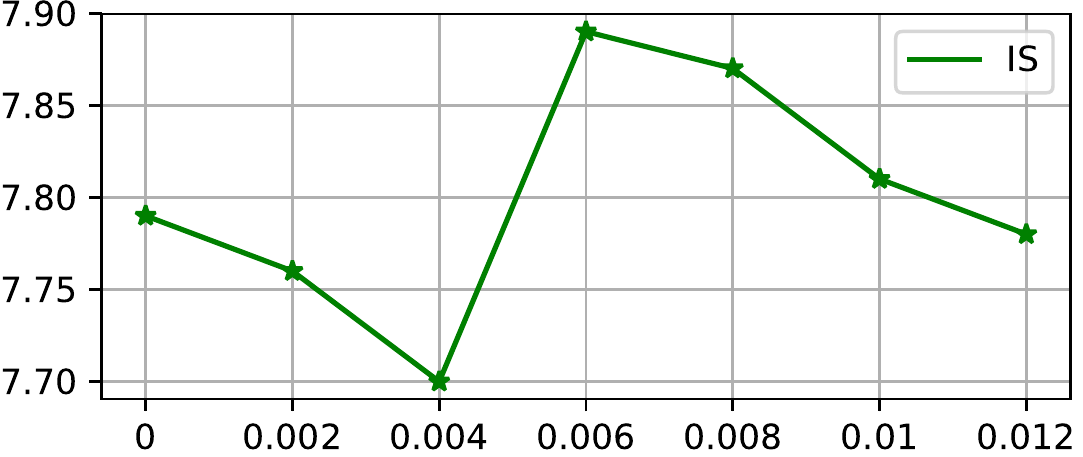}
  \caption{IS on CIFAR10}
  \label{fig:cmax_cifar10_IS}
\end{subfigure}
\begin{subfigure}{.5\textwidth}
  \centering
  \includegraphics[width=.9\linewidth]{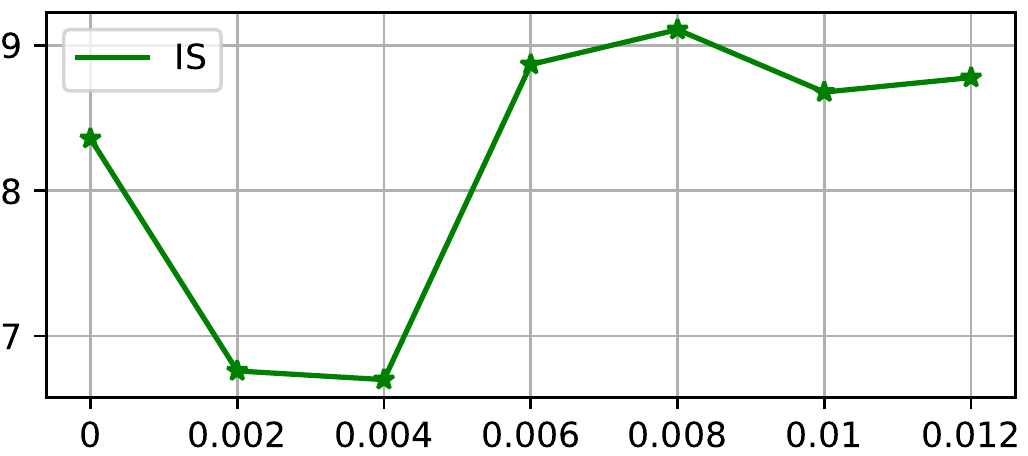}
  \caption{IS on CIFAR100}
  \label{fig:cmax_cifar100_IS}
\end{subfigure}
\caption{Generation quality under different $c_{\max}$ (x-axis) on CIFAR}
\label{fig:cmax_cifar}
\end{figure}

We try different values of $c_{\max}$ on CIFAR datasets, the results are shown in results in Figure~\ref{fig:cmax_cifar}.  Notice that due to the stochasticity in training process, the curves are \textbf{not} strictly monotone on both sides (a ``V''-shape). But we could see our final choice of $c_{\max}=0.006$ maximizes both CIFAR10 and CIFAR100 performances.

\subsection{Elongate training time on ImageNet-143}
Despite that our FastGAN reaches a better IS and FID than other models in shorter time, we would still like to see how it performs after fully convergence. To this end, we double the training iterations for FastGAN on ImageNet-143 experiments and present the results in Table~\ref{Tab:contimgnet143}. It turns out that FastGAN can further improve the IS and FID scores. Although the training time is doubled compared with the original runs, the total cost is still less than SNGAN and RobGAN. 
 \begin{table}[!htbp]
  \caption{Continue training results on ImageNet-143.}
  \label{Tab:contimgnet143}
  \centering
  \resizebox{0.95\textwidth}{!}{
  \begin{tabular}{lcccccc}
    \toprule
    & \multicolumn{3}{c}{64x64 pixels} & \multicolumn{3}{c}{128x128 pixels}  \\
    \cmidrule(r){2-4}\cmidrule(l){5-7}
 & IS $\uparrow$ & FID $\downarrow$ & Time $\downarrow$  & IS $\uparrow$ & FID $\downarrow$ & Time $\downarrow$ \\
    \midrule
Real data & $27.9\pm 0.42$  & $0.48$ & -- & $53.01\pm 0.56$  & $0.42$ & -- \\
\midrule
SNGAN & $10.70\pm 0.17$ &  $29.70$ & $2.2\times 10^5$ & $26.12\pm 0.29$ &  $30.83$ & $5.9\times 10^5$ \\
RobGAN & $\bm{24.62\pm0.31}$ & $14.64$ & $1.8\times 10^5$ & $33.81\pm 0.47$ & $33.98$ & $1.8\times 10^5$ \\
FastGAN & $22.23\pm 0.35$ & $12.04$  &  $\bm{3.6\times 10^4}$ & $40.41\pm 0.49$ &  $14.48$ & $\bm{7.1\times 10^4}$ \\
FastGAN(cont.) & $\bm{23.93\pm 0.26}$ & $\bm{10.54}$  &  $7.2\times 10^4$ & $\bm{42.91\pm0.57}$ &  $\bm{11.98}$ & $1.4\times 10^5$ \\

    \bottomrule
  \end{tabular}
  }
\end{table}

\section{More samples to compare the loss}
\begin{figure}[!h]
\begin{subfigure}{.3\textwidth}
  \centering
  \includegraphics[width=.9\linewidth]{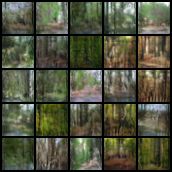}
    \includegraphics[width=.9\linewidth]{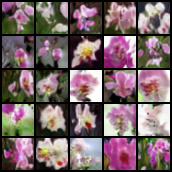}
    \includegraphics[width=.9\linewidth]{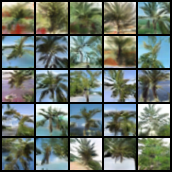}
    \includegraphics[width=.9\linewidth]{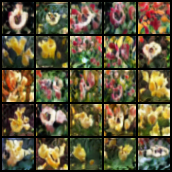}
    \includegraphics[width=.9\linewidth]{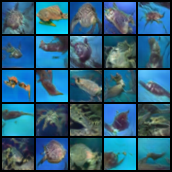}
  \caption{\centering{Improved loss \newline (IS:8.87, FID:17,27)}}
  \label{fig:cifar100_newloss}
\end{subfigure}
\begin{subfigure}{.3\textwidth}
\centering
    \includegraphics[width=.9\linewidth]{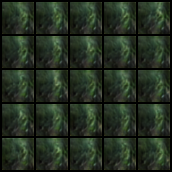}
    \includegraphics[width=.9\linewidth]{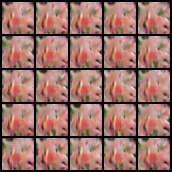}
    \includegraphics[width=.9\linewidth]{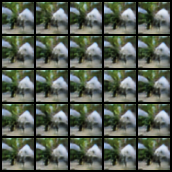}
    \includegraphics[width=.9\linewidth]{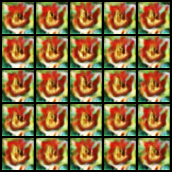}
    \includegraphics[width=.9\linewidth]{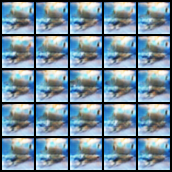}
  \caption{\centering{RobGAN loss \newline (IS:6:26, FID:43.69)}}
  \label{fig:cifar100_robganloss}
\end{subfigure}
\begin{subfigure}{.3\textwidth}
  \centering
  \includegraphics[width=.9\linewidth]{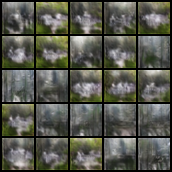}
  \includegraphics[width=.9\linewidth]{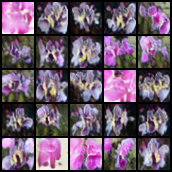}
    \includegraphics[width=.9\linewidth]{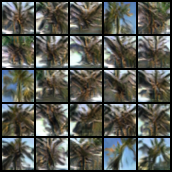}
    \includegraphics[width=.9\linewidth]{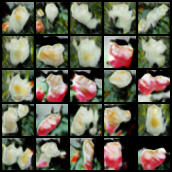}
    \includegraphics[width=.9\linewidth]{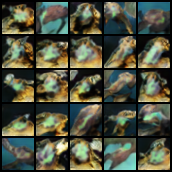}
  \caption{\centering{Improved loss w/o KL-term \newline (IS:7.42, FID:24.02)}}
  \label{fig:cifar100_newlosswoKL}
\end{subfigure}
\label{fig:cifar100_losscompare}
\caption{Images randomly generated from FastGAN models trained on CIFAR100 via using different losses. Each row is a class. The new proposed improved loss format (a) is able to generate better images with higher intra-class diversity. RobGAN loss (b) collapses inside each classes.}
\end{figure}

\begin{figure}[!h]
\centering
\begin{subfigure}{.3\textwidth}
  \includegraphics[width=.9\linewidth]{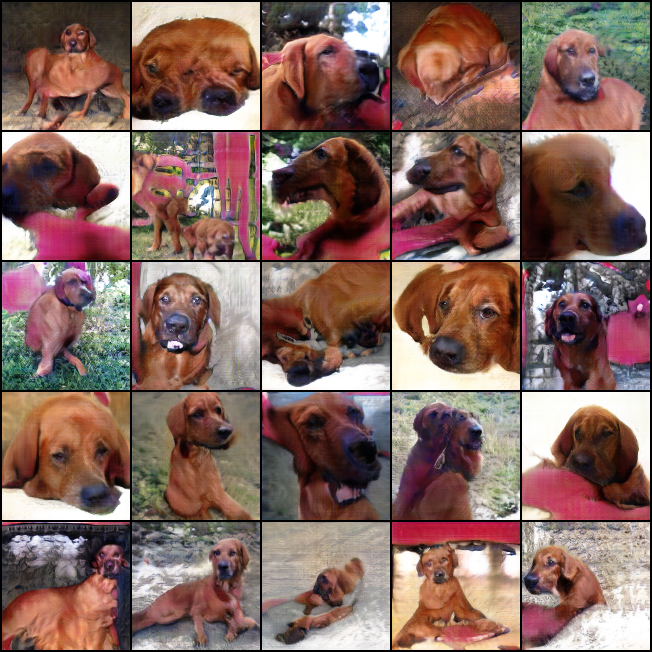}
    \includegraphics[width=.9\linewidth]{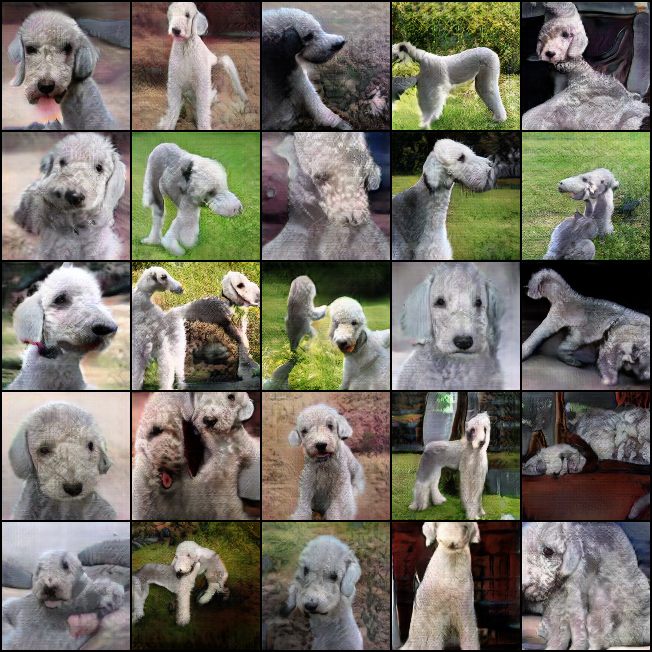}
        \includegraphics[width=.9\linewidth]{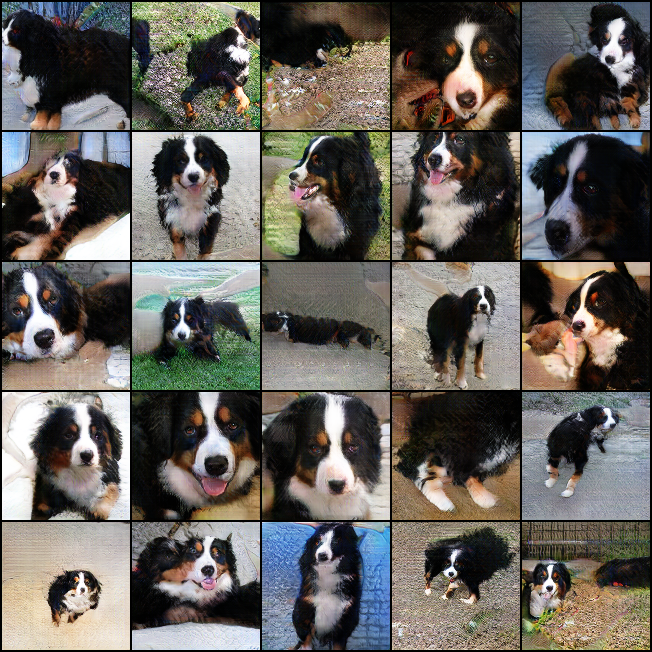}
            \includegraphics[width=.9\linewidth]{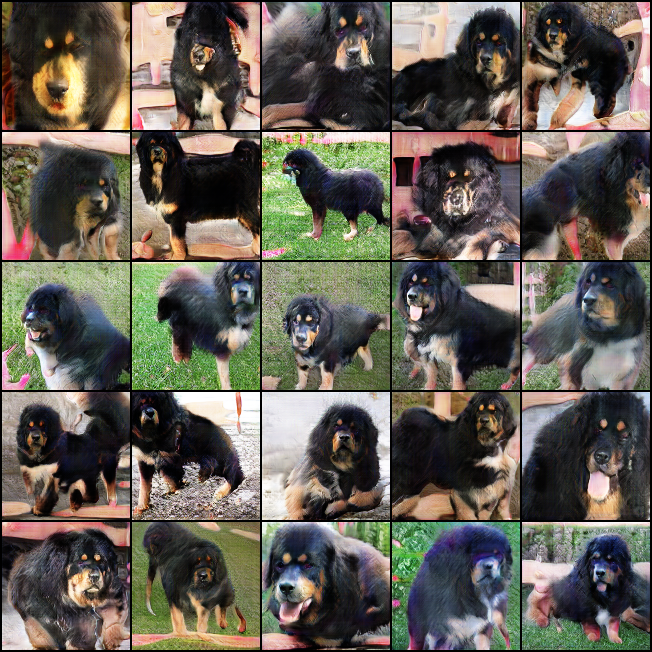}
  \caption{\centering{Improved loss \newline (IS:40.41, FID:14.48)}}
  \label{fig:catdog128_newloss}
\end{subfigure}%
\begin{subfigure}{.3\textwidth}
  \centering
  \includegraphics[width=.9\linewidth]{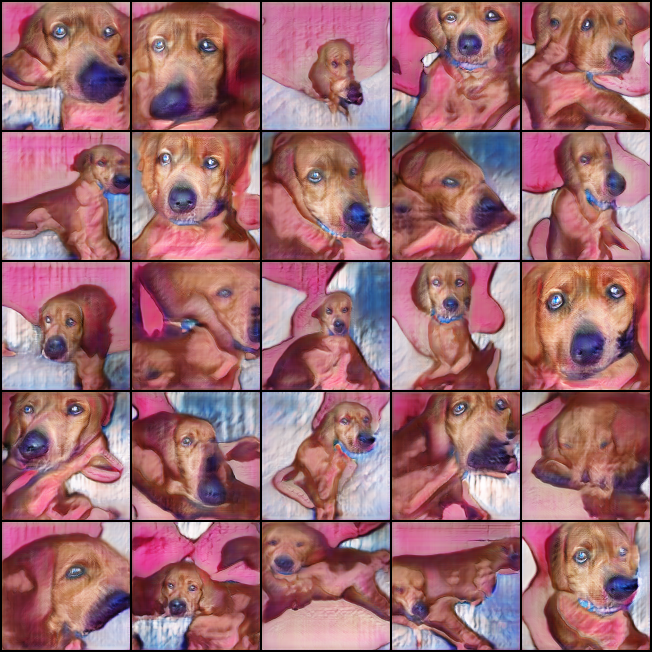}
    \includegraphics[width=.9\linewidth]{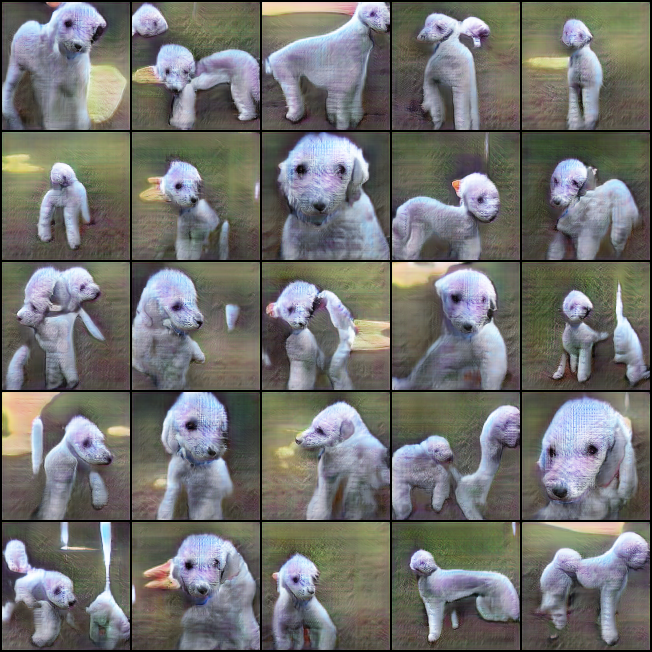}
        \includegraphics[width=.9\linewidth]{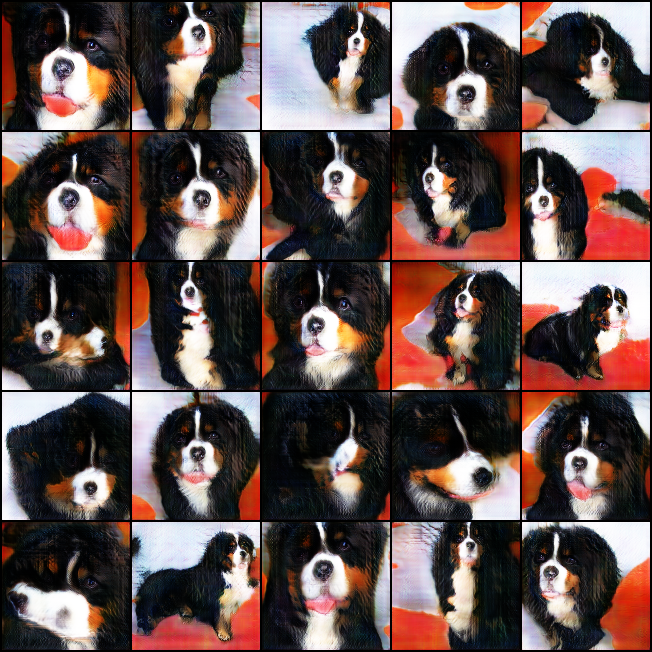}
            \includegraphics[width=.9\linewidth]{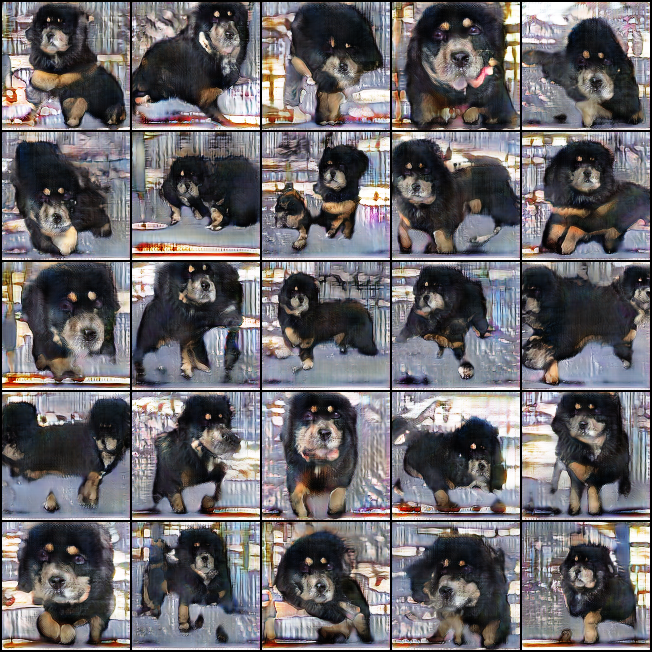}
  \caption{\centering{RobGAN loss \newline (IS:45.94, FID:25.38)}}
  \label{fig:catdog128_robganloss}
\end{subfigure}
\begin{subfigure}{.3\textwidth}
  \centering
  \includegraphics[width=.9\linewidth]{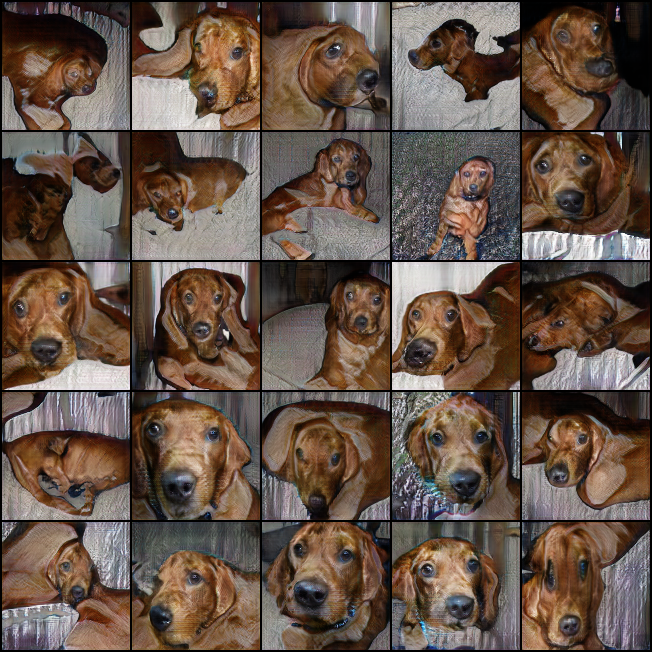}
  \includegraphics[width=.9\linewidth]{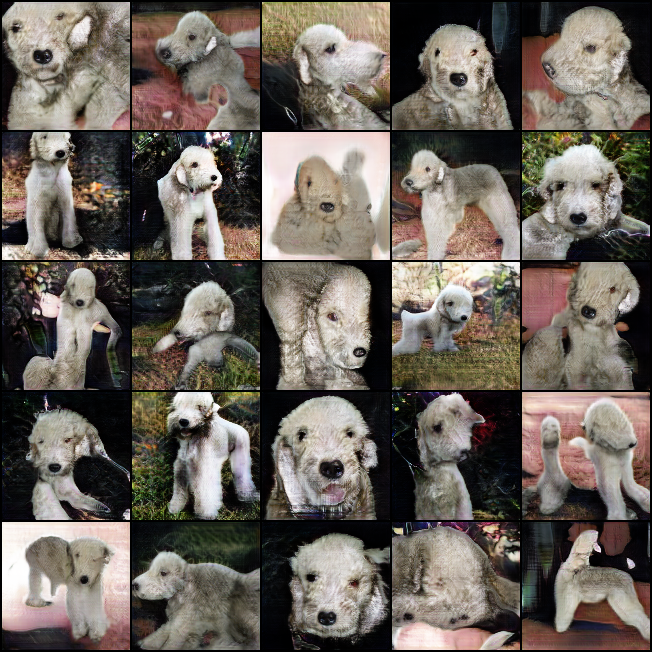}
    \includegraphics[width=.9\linewidth]{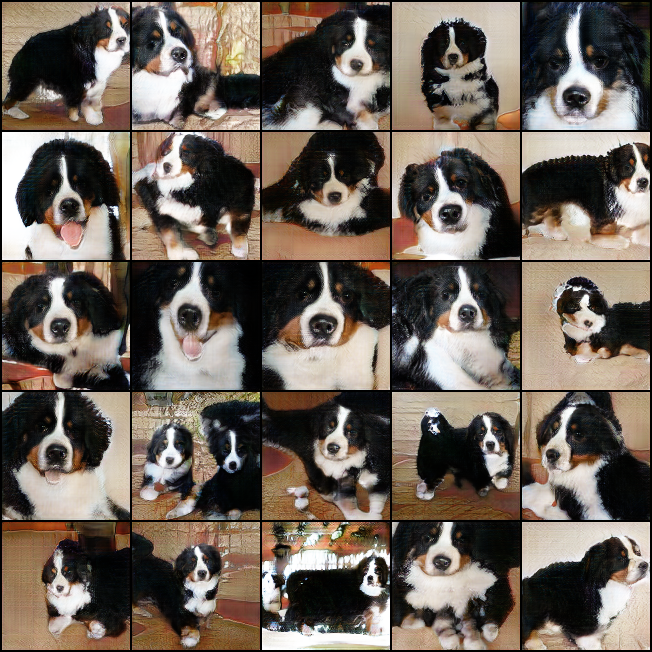}
        \includegraphics[width=.9\linewidth]{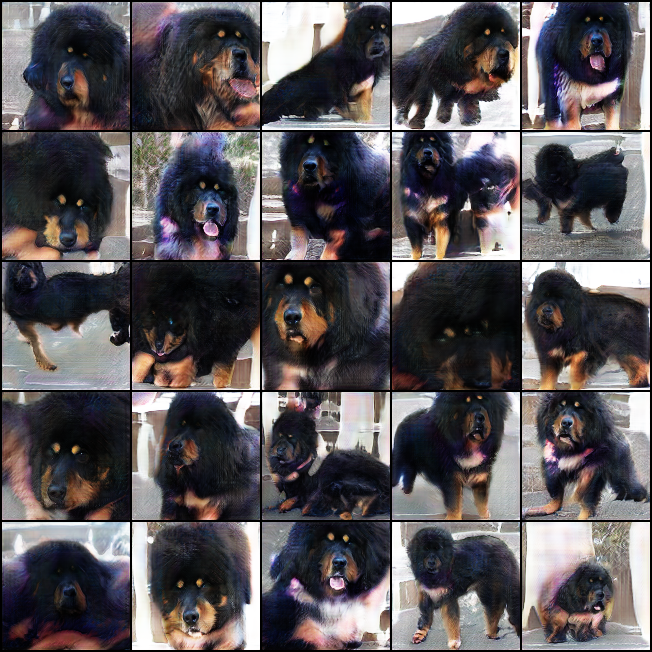}
  \caption{\centering{Improved loss w/o KL-term \newline (IS:42.54, FID:17.51)}}
  \label{fig:catdog128_newlosswoKL}
\end{subfigure}
\label{fig:catdog128_losscompare}
\caption{Images randomly generated from FastGAN models trained on ImageNet-143 (128 $\times$ 128) via using different losses. Each row is a class. The new proposed improved loss format (a) is able to generate better images with higher intra-class diversity.}

\section{Additional samples on ImageNet}
\end{figure}
\begin{figure}[!h]
\begin{center}
\includegraphics[width=1.0\linewidth]{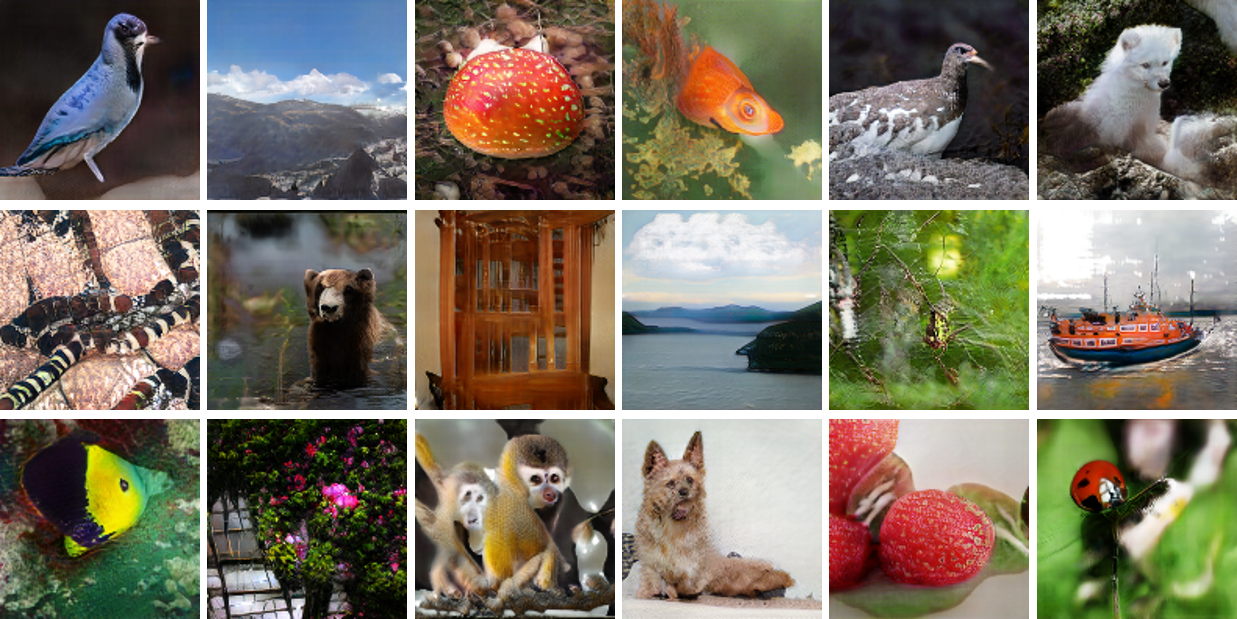}
\end{center}
\vspace{-7pt}
   \caption{Images randomly sampled from FastGAN trained on ImageNet with batch size 64}
\label{fig:imgnet_sample_64}
\end{figure}

\begin{figure}[!h]
\begin{center}
\includegraphics[width=1.0\linewidth]{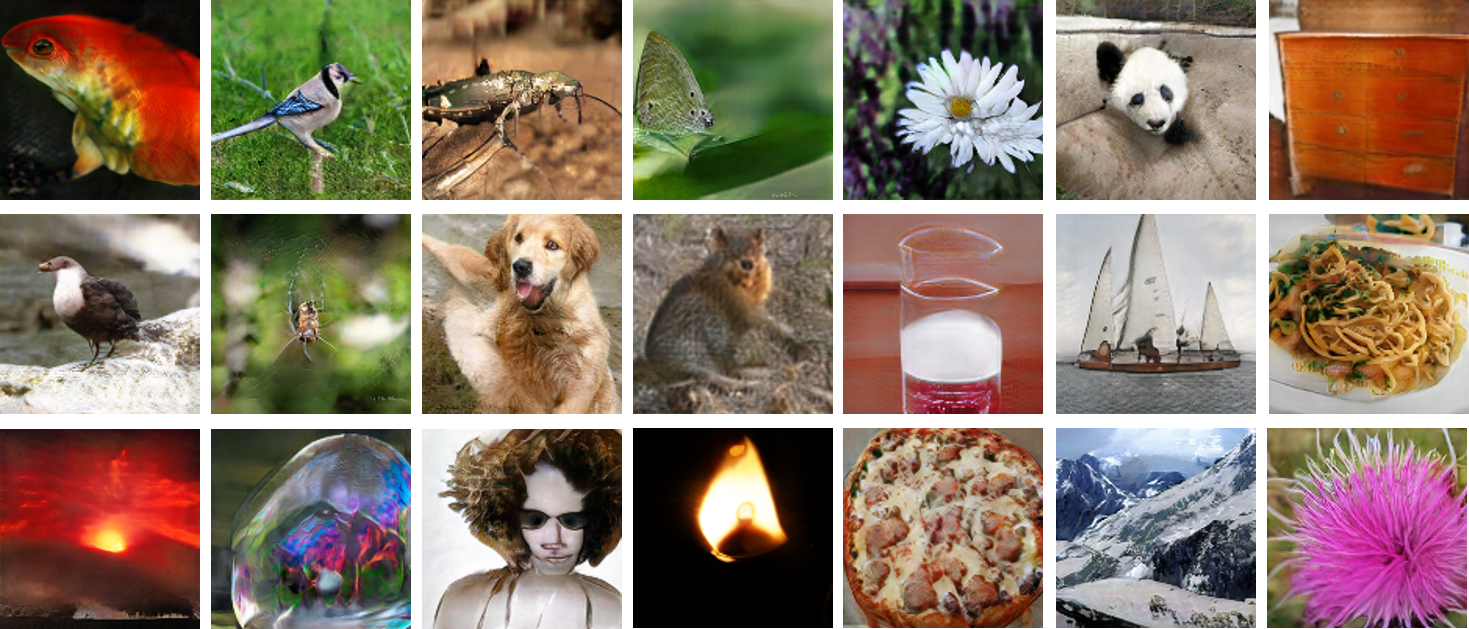}
\end{center}
\vspace{-7pt}
   \caption{Images randomly sampled from FastGAN trained on ImageNet with batch size 256}
\label{fig:imgnet_sample_256}
\end{figure}

\clearpage

{\small
\bibliographystyle{ieee}
\bibliography{refs}
}

\end{document}